\documentclass{article}

\usepackage{arxiv}

\usepackage[utf8]{inputenc} % allow utf-8 input
\usepackage[T1]{fontenc}    % use 8-bit T1 fonts
\usepackage{hyperref}       % hyperlinks
\usepackage{url}            % simple URL typesetting
\usepackage{booktabs}       % professional-quality tables
\usepackage{amsfonts}       % blackboard math symbols
\usepackage{nicefrac}       % compact symbols for 1/2, etc.
\usepackage{microtype}      % microtypography
\usepackage{lipsum}         % Can be removed after putting your text content
\usepackage{graphicx}
\usepackage{natbib}
\usepackage{doi}
\usepackage{algorithm}
\usepackage{algorithmic}
\usepackage{amsmath}
\usepackage{array}
\usepackage{multirow}
\usepackage{color}
\usepackage{cleveref} 
\usepackage{subfigure} 
\usepackage{enumitem}
\usepackage{natbib}
\setcitestyle{numbers,square}

\title{Video RWKV:Video Action Recognition based RWKV }

\newif\ifuniqueAffiliation
\uniqueAffiliationtrue

\ifuniqueAffiliation % Standard variant of author block
\author{{\includegraphics[scale=0.06]{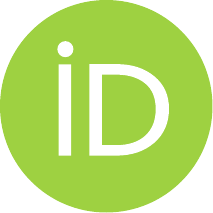}\hspace{1mm}Zhuowen Yin}
	\\
	Northeastern University\\
	Shenyang \\
	\\
	\And
	{\includegraphics[scale=0.06]{orcid.pdf}\hspace{1mm}Chengru Li} \\
	\\
	Anhui University\\
	Hefei \\
	 \\
	\And
	{\includegraphics[scale=0.06]{orcid.pdf}\hspace{1mm}Xingbo Dong} \\
	\\
	Anhui University\\
	Hefei \\
	 \\
}
\else
\usepackage{authblk}

\newbox{\orcid}\sbox{\orcid}{\includegraphics[scale=0.06]{orcid.pdf}} 
\author[1]{%
	\href{https://orcid.org/0000-0000-0000-0000}{\usebox{\orcid}\hspace{1mm}David S.~Hippocampus\thanks{\texttt{hippo@cs.cranberry-lemon.edu}}}%
}
\author[1,2]{%
	\href{https://orcid.org/0000-0000-0000-0000}{\usebox{\orcid}\hspace{1mm}Elias D.~Striatum\thanks{\texttt{stariate@ee.mount-sheikh.edu}}}%
}
\affil[1]{Department of Computer Science, Cranberry-Lemon University, Pittsburgh, PA 15213}
\affil[2]{Department of Electrical Engineering, Mount-Sheikh University, Santa Narimana, Levand}
\fi

\hypersetup{
pdftitle={A template for the arxiv style},
pdfsubject={q-bio.NC, q-bio.QM},
pdfauthor={David S.~Hippocampus, Elias D.~Striatum},
pdfkeywords={First keyword, Second keyword, More},
}

\begin{document}
\maketitle
\begin{abstract}
    \hspace{1em}To address the challenges of high computational costs and long-distance dependencies in existing video understanding methods, such as CNNs and Transformers, this work introduces RWKV to the video domain in a novel way. We propose a LSTM CrossRWKV (LCR) framework, designed for spatiotemporal representation learning to tackle the video understanding task. Specifically, the proposed linear complexity LCR incorporates a novel Cross RWKV gate to facilitate interaction between current frame edge information and past features, enhancing the focus on the subject through edge features and globally aggregating inter-frame features over time. LCR stores long-term memory for video processing through an enhanced LSTM recurrent execution mechanism. By leveraging the Cross RWKV gate and recurrent execution, LCR effectively captures both spatial and temporal features. Additionally, the edge information serves as a forgetting gate for LSTM, guiding long-term memory management.Tube masking strategy reduces redundant information in food and reduces overfitting.These advantages enable LSTM CrossRWKV to set a new benchmark in video understanding, offering a scalable and efficient solution for comprehensive video analysis. All code and models are publicly available.
\end{abstract}    
\section{Introduction}
\label{sec:intro}

\hspace{1em} The growing prominence of video understanding is underscored by the rapid rise of short video platforms. The primary objective in this domain is to effectively capture spatio-temporal features. However, video often contain significant redundant information, posing challenges for efficient processing. Most existing approaches rely on 3D-CNNs or Transformer-based architectures to extract spatio-temporal features, either through local convolutions or long-range attention mechanisms. These methods require significant computational resources, which leads to limitations in their scalability and practical deployment.\cite{carreira2017quo,feichtenhofer2020x3d,feichtenhofer2019slowfast,hara2018can,ji20123d,xie2018rethinking}

\hspace{1em} Recently, traditional RNN models and state space models (SSMs) have garnered significant attention due to their ability to capture long-term information while maintaining linear time complexity. Mamba\cite{gu2023mamba} enhances SSMs by incorporating time-varying parameters and introduces a hardware-aware algorithm for highly efficient training and inference. RWKV\cite{peng2023rwkv,peng2024eagle} improves the linear attention mechanism, addressing the parallelism challenges of RNNs, and achieves RNN-like time complexity with performance comparable to Transformers. xLSTM\cite{beck2024xlstm}, by introducing exponential gating and an enhanced memory structure, overcomes the limitations of traditional LSTMs.The success of these models in natural language processing has inspired their application to vision domain, such as Vision Mamba\cite{zhu2024vision} and Vision RWKV\cite{duan2024vision}. However, these models struggle to efficiently capture the temporal dynamics of long video sequences.

\begin{figure}[!h] % figure环境 % [控制图片放置的位置]
  \centering % 图片居中显示
  \includegraphics[width=1.0\columnwidth]{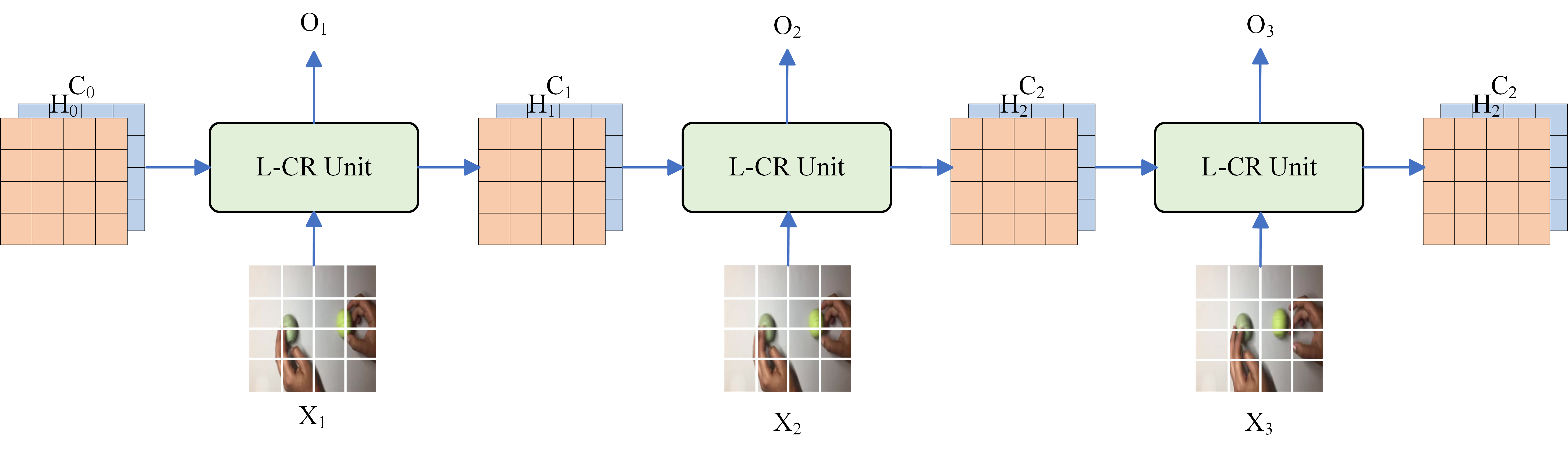} % {图片文件的名字}
  \caption{The overview of the LSTM CrossRWKV working pipeline.The input frame $X_{t}$, hidden state $H_{t}$ and cell state $C_{t}$ jointly determine output $O_{t}$, The hidden state and cell state contains memory information which can be transferred to next frame.
  } 
  \label{tu2}
 \end{figure}

 \hspace{1em} Inspired by this, we designed LSTM-Cross RWKV(LCR), a model specifically tailored for video understanding. To further enhance temporal data representation, LSTM\cite{6795963} is employed to capture long sequence information.CrossRWKV integrates past temporal information with the current frame's edge prompting data, offering a linear complexity approach for dynamic spatio-temporal context modeling. By utilizing edge prompt, the model significantly reduces redundant information in video sequences, leading to more efficient and accurate video understanding.\\
 \textbf{Our contributions are as follows:}
\begin{enumerate}[label=\roman*.]
  \item We propose the LSTM-CrossRWKV framework for processing video sequences in video understanding task, an innovative recurrent cell that fuses an LSTM architecture with Cross RWKV Blocks.which is able to efficiently extract spatiotemporal representations.
  \item We introduce a novel CrossRWKV  into the LSTM-CrossRWKV unit, which preserves a blend of past and current temporal information via the receiver vector, while the key and value components incorporate the edge information of the current frame.
  \item We evaluate the effectiveness of the proposed model on Kinetics-400, Sometingsometing-V2, and Jester.Experimental results show that LSTM-CrossRWKV achieves
  excellent performance on three datasets.
\end{enumerate}

\section{Related Work}
\label{sec:formatting}

\textbf{Convolutional neural networks (CNNs)} Convolutional neural networks (CNNs) have shown remarkable performance in computer vision tasks, with methods based on both 2D CNNs\cite{carreira2017quo,feichtenhofer2019slowfast,feichtenhofer2016convolutional,simonyan2014two,tran2018closer} and 3D convolution\cite{fan2019more,karpathy2014large,kwon2020motionsqueeze,qiu2017learning,wang2018non,hara2018can}. In video understanding tasks, combining CNNs with RNNs effectively captures the nuances of spatio-temporal relationships, thereby improving prediction accuracy.ConvLSTM\cite{shi2015convolutional} enhances traditional LSTM by incorporating convolutional operations in place of fully connected layers, allowing it to better capture spatial features in the input.PredRNN\cite{wang2022predrnn} enables memory states across different LSTM layers to pass information both vertically and horizontally, while PredRNN++\cite{wang2018predrnn++} addresses the issue of gradient vanishing in deep RNN models with a Gradient Highway unit. E3D-LSTM\cite{wang2018eidetic} augments ST-LSTM's memory capacity using 3D convolutions, and the MIM model redesigns the forgetting gate with dual recursive units to better handle the learning of non-stationary information. CrevNet\cite{yu2019crevnet} introduces a reversible CNN-based architecture for decoding complex spatio-temporal patterns, and PhyDNet\cite{guen2020disentangling} leverages CNN simulation bias to integrate prior physical knowledge, improving prediction quality.In summary, these models demonstrate diverse approaches to enhance the capture of spatio-temporal dependencies with remarkable results. However, traditional convolutional methods tend to focus too heavily on local information, often overlooking broader spatio-temporal dependencies in video data.\\
\textbf{Transformer} Transformers were initially proposed for natural language\cite{devlin2018bert,vaswani2017attention} processing and achieved remarkable success, leading to their exploration in computer vision. The Vision Transformer (ViT)\cite{dosovitskiy2020image} demonstrated that a pure Transformer architecture can perform well in image classification tasks. The end-to-end model DERT\cite{carion2020end} further enhances this by integrating convolutional operations for object detection. The Swin Transformer\cite{liu2021swin} has made significant advancements with its innovative shift-window strategy and hierarchical structure, resulting in impressive performance across various tasks. Building on this foundation, SwinLSTM\cite{tang2023swinlstm} innovatively combines Swin Transformer with LSTM, establishing a robust benchmark for spatio-temporal prediction. ViViT\cite{arnab2021vivit} addresses this by employing two Transformer encoders to separately extract spatial and temporal features. RViT\cite{yang2022recurring} combines the strengths of both RNNs and ViT, utilizing a frame-stream processing technique that conserves GPU memory. Timesformer\cite{bertasius2021space} introduces divided space-time attention to efficiently reduce computational effort while enhancing results. Despite these advancements, many existing Transformer-based methods struggle to efficiently handle high-resolution or long-duration videos due to the quadratic time complexity of the original attention mechanism, thus subsequent approaches focus on reducing the time complexity of the attention mechanism to improve efficiency.Longformer\cite{beltagy2020longformer} enables processing of lengthy sequences by incorporating an attention mechanism with O(n) complexity, which combines local contextual self-attention and task-specific global attention. VTN\cite{neimark2021video} builds an efficient architecture for video comprehension by utilizing a 2D spatial feature extraction model alongside a temporal attention-based encoder. MViT\cite{fan2021multiscale} proposes multi-head pooled attention with a specific spatial-temporal resolution, effectively reducing input sequence length and achieving encouraging results.XciT\cite{ali2021xcit} reduces the quadratic complexity of traditional attention computation to linear complexity using the covariance matrices of Q and K.\\
\hspace{1em} Based on the above research, we aim to fuse LSTM with Cross RWKV. By utilizing the edge information of the current frame as a prompt, we can effectively integrate it into the network to reduce the impact of redundant information. Cross RWKV will be employed to merge past temporal information and the current frame edge information, enhancing the model's focus on relevant subject information.The proposed LSTMRWKV network is specifically designed for video understanding tasks.

\section{LSTM-Based RWKV}

\hspace{1em}Previous work revolves around CNN and ViT for spatiotemporal features extraction, the quadratic linear complexity hinders the efficient operation of the algorithm. To address this, we propose an LSTM-CrossRWKV model based Vision RWKV and our LSTM effectively captures spatiotemporal features from the video. The edge information is utilized through a unique gating mechanism that serves as prompt information to guide the model's attention. By employing the Cross RWKV gate to aggregate past and current frame edge information, our model intelligently integrates multimodal information. This approach enhances task accuracy and robustness.

\hspace{1em} In the following subsection, we will first discuss the patch embedding of each frame in the pre-processing stage of LSTM-CrossRWKV. The specifically designed Cross RWKV gate then is introductioned. The processing pipeline of the LCR unit is futher discussed, followed by the Edge Prompt Learning. The whole framework will be presented finally.

\begin{figure}[htbp] % figure环境 % [控制图片放置的位置]
 \centering % 图片居中显示
 \includegraphics[width=1.0\columnwidth]{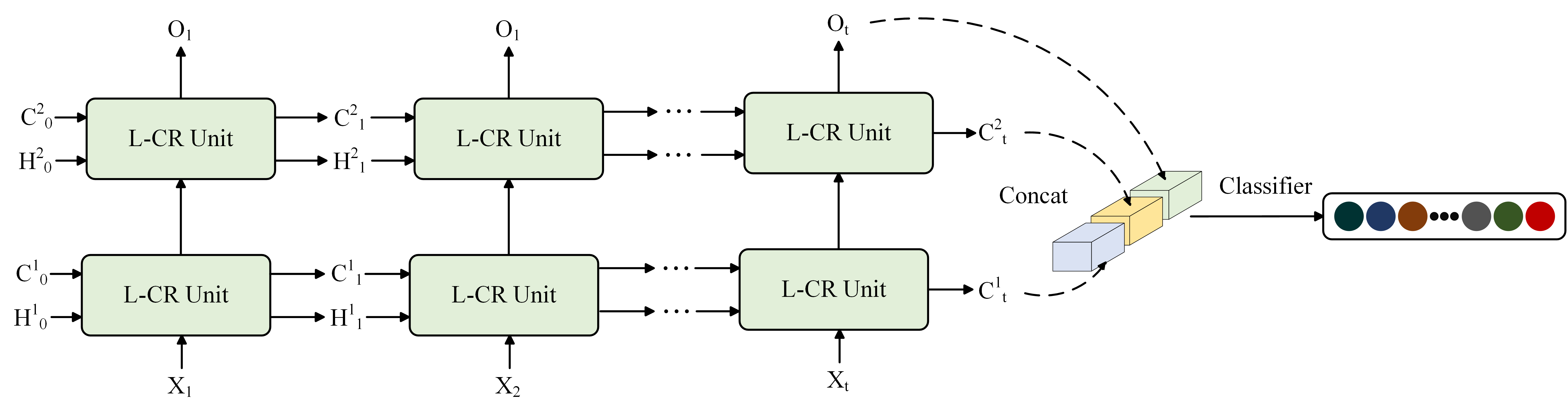} % {图片文件的名字}
 \caption{The framework of LSTM CrossRWKV, we use frame-by-frame analysis for video recognition, which effectively reduces the memory footprint and improves the inference speed
 }  % 给图片添加标题
 \label{tu2}
\end{figure}

%-------------------------------------------------------------------------
\subsection{LSTM-CrossRWKV}
\hspace{1em}In the pre-processing stage, we use 3D convolution (i.e., 1$\times$16$\times$16) to divide the input video $X^v\in\mathbb{R}^{3\times T\times H\times W}$ into L non-overlapping spatiotemporal patches $x\in\mathbb{R}^{L\times C}$,where $L{=}t{\times}h{\times}w (t{=}T, h{=}\frac{H}{P}$ and $ w{=}\frac{W}{P})$.The sequence of tokens input to the following LSTM-Cross RWKV encoder is
\begin{equation}
  \label{}
  x = \mathrm{Concat}(\ell(X^{v}),X_{cls}) + Pos
\end{equation}
where $\ell$ is an embedding function (a 3D convolution layer in our work),$X_{cls}$ is a learnable classification token that is appended to the last dimension of the feature sequence. After feature aggregation, $X_{cls}$ contains global information, which is then processed by normalization and a linear layer for classification tasks. We introduced $Pos\in\mathbb{R}^{P^2\times C}$ a learnable rotary position encoding to better capture the relative positional information between video frames. The tokens $X^{v}$ are then passed through L LSTM-CrossRWKV blocks, and the representation of the [CLS] token at the final layer is processed by normalization and a linear layer for classification.

%-------------------------------------------------------------------------
\subsection{Cross RWKV Gate}
\hspace{1em}Given the current video frame $x_{t}$ and the corresponding edge image $x_{t}^e$, along with the hidden state $h_{t-1}\in\mathbb{R}^{P^2\times C}$ and the cell state $c_{t-1}\in\mathbb{R}^{P^2\times C}$from the LSTM, the inputs are first aggregated using causal convolution. In each layer, the input is initially processed by the spatial mixing module, which serves as a global attention mechanism. Here, both the current image and edge information are shifted and passed through three parallel linear layers to produce a matrix. 
\begin{equation}
  \begin{split}
   \label{}
   R=\mathrm {Q\text{-}Shift_{R}}(Cauconv(x_{t},h_{t-1}))W_{R}\\
   G=\mathrm {Q\text{-}Shift_{G}}(Cauconv(x_{t},h_{t-1}))W_{G}
   \end{split}
 \end{equation}
 \begin{equation}
  \begin{split}
   \label{}
   K=\mathrm {Q\text{-}Shift_{K}}(x_{t}^e)W_{K}\\
   V=\mathrm {Q\text{-}Shift_{V}}(x_{t}^e)W_{V}
  \end{split}
 \end{equation}
Here,$K$ and $V$ are used to compute the cross attention results through a linear complexity bidirectional attention mechanism.
\begin{equation}
  a_t=\mathrm{concat}\big(\mathrm{SiLU}(\mathrm{G}_t)\odot\mathrm{LN}(R_t\cdot Bi\text{-}{w}KV_t)\big)\boldsymbol{W}_a
  \end{equation}
  \begin{equation}
    wKV_t=\mathrm{diag}(u)\cdot K_t^\mathrm{T}\cdot V_t+\sum_{i=1}^{t-1}\mathrm{diag}(w)^{t-1-i}\cdot K_i^\mathrm{T}\cdot V_i
    \end{equation}
where LN indicates the LayerNorm, it operates on each of $h$ heads separately. It is also worth noting that $w$ is obtained from $w=\exp(-\exp(\omega))$, where $\omega\in\mathbb{R}^{D/h}$ are the actual headwise trainable parameters.\\ 
\hspace{1em} Subsequently, the tokens are passed into the channel-mix module for a channel
wise fusion. $R_{t}^{\prime}$, $K_{t}^{\prime}$ are obtained in a similar manner as spatial-mix:
\begin{equation}
  \begin{split}
 R_{t}^{\prime} =\mathrm{Q\text{-}Shift_{R^{\prime}}}(a_{t})W_{R^{\prime}} \\
 K_{t}^{\prime} =\mathrm{Q\text{-}Shift_{R^{\prime}}}(a_{t})W_{K^{\prime}} \\
  \end{split}
 \end{equation}
 Here, $Vc$ is a linear projection of $K$ after the activation function and the output
 $A_{t}^{\prime}$ is also controlled by a gate mechanism $\sigma(R_{t}^{\prime})$ before the output projection:
 \begin{equation}
  V_{t}^{\prime} = \mathrm{ReLU}(K_{t}^{\prime})^{2}W_{V^{\prime}}
\end{equation}
\begin{equation}
  A_{t}^{\prime} = \sigma(R_{t}^{\prime}) \odot V_{t}^{\prime}
\end{equation}
\hspace{1em}  Simultaneously, residual connections [20] are established from the tokens to each normalization layer to ensure that training gradients do not vanish in deep networks.A diagram of the attention gate is shown in Figure \textcolor{red}{3}.
\begin{figure}[htbp]
  \centering
  \begin{minipage}[t]{0.369\textwidth}
    \includegraphics[width=\textwidth]{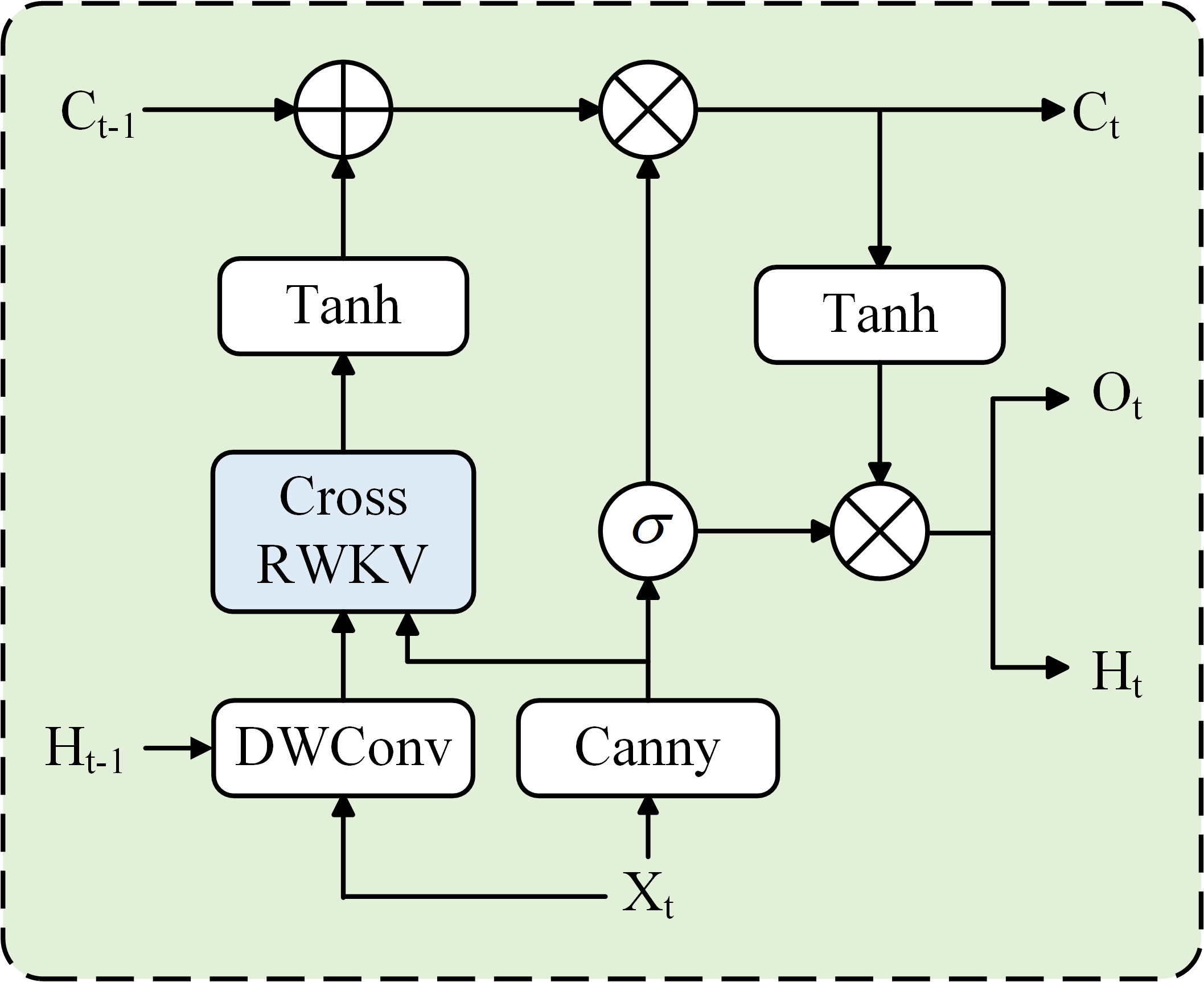}
    \centering (a). LCRUnit
    \label{fig:sourceImage}
  \end{minipage}
\hfill
  \begin{minipage}[t]{0.623\textwidth}
    \centering
    \includegraphics[width=\textwidth]{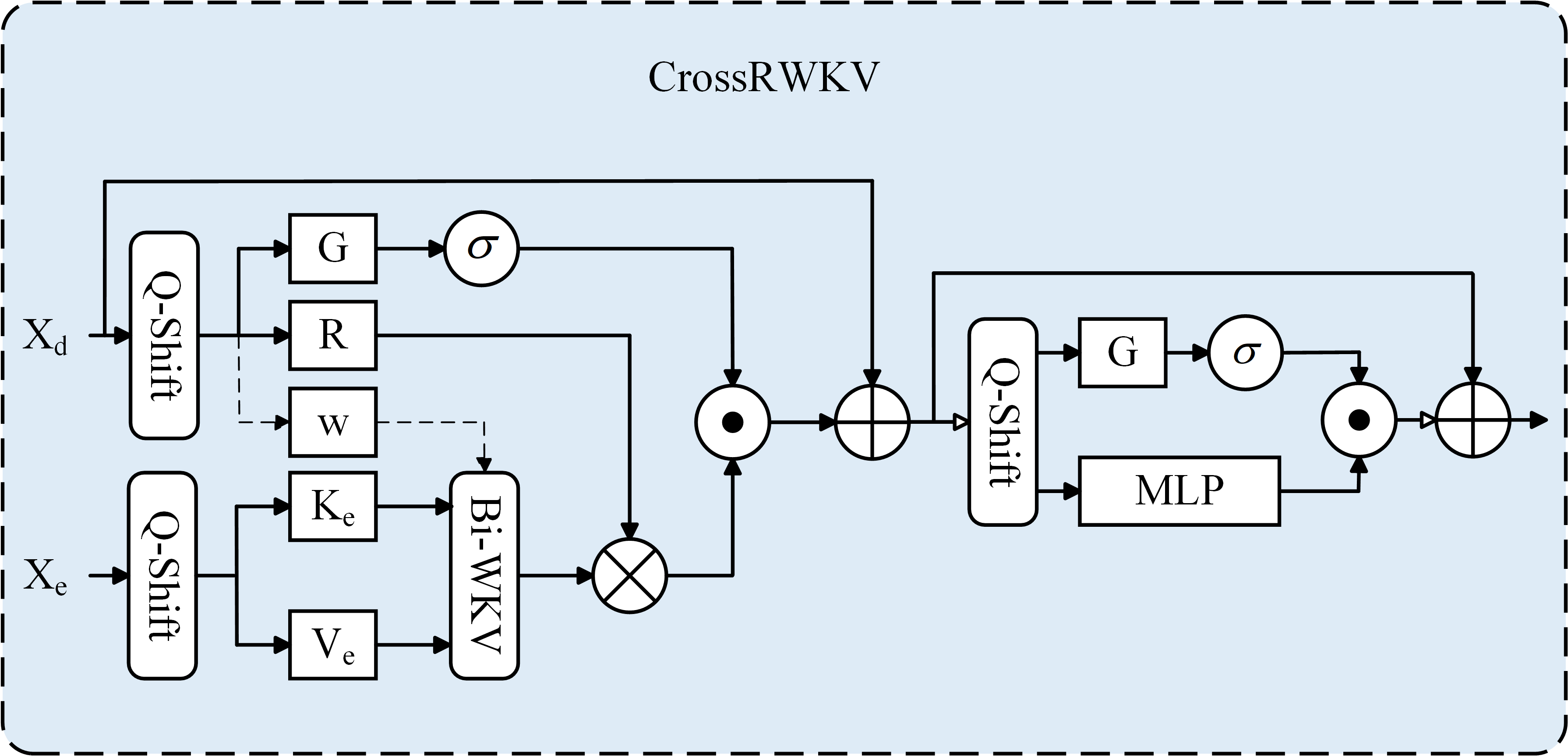}
    \\(b). CrossRWKV
    \label{fig:groundTruth}
  \end{minipage}
\caption{LCR Figure (3a) shows a recurrent unit in our framework. Figure (3b) illstrastes how Cross RWKV process both current input $X_{d}$ and current edge feature $X_{e}$.}
\label{fig:result}
\end{figure}

%-------------------------------------------------------------------------
\subsection{Edge Prompt Learning}
\hspace{1em}The current frame $x_{t}$ is subjected to edge extraction by the $Canny$ operator, and the two thresholds are adaptively determined using $Otsu's$ method.
\begin{equation}
  x_{t-1}^e=\text{Z}(\mathrm{Canny}(x_t))
\end{equation}
where $\text{Z}$ indicates the zero embedding function(zero-initialized convolution layers), since both the weight and bias parameters of a zero convolution layer are initialized to zero.The edge information is passed through the forgetting gate of LSTM to update the cell state, which is used to reduce the effect of background noise.

%-------------------------------------------------------------------------
\subsection{LCR Unit}
\hspace{1em} An overview diagram of a single LCR unit is shown in Figure \textcolor{red}{2a}. Firstly, $x_{t}$ and $h_{h-1}$ are passed by the caucal convolution layer and CrossRWKV gate. Next $Ct$, $Ht$ and $Ot$ are obtained through input gates, forget gates and output gates,and finally we concat $X_{cls}^O$ and $X_{cls}^C$ are transferred to the linear layer $W_{class}$ for classification. LCR unit consist of three key equations.
\begin{equation}
  C_{t}=x_t^e \times (\mathrm{Tanh}(A_{t}) + C_{t-1})+C_{t-1}
\end{equation}
\begin{equation}
  O_{t}=H_{t}=\operatorname{Tanh}(C_{t})\times\sigma(A_t)
\end{equation}
\begin{equation}
  result=W_{class}\times\mathrm{LN}(Concat(X_{cls}^O,X_{cls}^C))
\end{equation}
where $\sigma$ indicates the activation function $Sigmoid(\cdot)$ and LN indicates the LayreNorm. 

%-------------------------------------------------------------------------
\subsection{LSTM-CrossRWKV Framework}
\hspace{1em}Existing methods typically process a batch of frames. For example, 3D-ResNet and TimeSformer require relatively long video sequences for inference and training. In contrast, our method processes video frame-by-frame; we use a tube masking strategy to eliminate redundant information in the video; We add edge information as a prompt to enhance the network's attention to the subject and reduce the effect of noise on the network by exploiting the LSTM's forgetting gate and Cross RWKV.

\section{Experiments and Results}

%-------------------------------------------------------------------------
\subsection{Implement Details}
\textbf{Datasets} To evaluate the proposed method, we use three public benchmark datasets for human action recognition tasks, including Kinetics-400 (K400)\cite{carreira2017quo,kay2017kinetics} ($\sim$240K training videos and $\sim$20K validation videos from 400 human action categories), Jester\cite{materzynska2019jester} (120K training videos from 27 human gestures), Something- Something V2(SSv2)\cite{goyal2017something} ($\sim$168.9K training videos and $\sim$24.7K validation videos from 174 categories).\\
\textbf{Training} For the Kinetics-400 dataset, we first resize each video to 256 $\times$ 256, then extract a clip from the full-length video. The clip is randomly cropped to 224 $\times$ 224 and horizontally flipped at random. Since our architecture is based on Vision RWKV (VRWKV), we initialize the model with an ImageNet-1K pre-trained VRWKV for the Kinetics-400 experiments. The preprocessing pipeline for the SSv2 dataset is identical. Jester dataset, where the video length may not suffice to sample 32 frames, we pad the video by randomly duplicating frames. All frames are resized to 112x112 pixels without additional transformations, and we train the model from scratch on the Jester dataset. The training process employs label smoothing cross-entropy loss.\\
\hspace{1em} Top-1 and Top-5 accuracy $(\%)$ are used for evaluation on each validation dataset. Additionally, we include the total model parameters, computation cost (FLOPs), and memory consumption for one-view inference in the subsequent experiments. Notably, we utilize the official code \cite{arnab2021vivit,contributors2020openmmlab,fan2020pyslowfast,fan2021pytorchvideo,fan2019more,lin2019tsm,patrick2021keeping,tran2019video} (if available) to verify other methods. Various model settings are designed to assess the performance of our framework under different conditions. Detailed configurations for each dataset are presented in Table \textcolor{red}{1}.\\
\begin{table}[H]
    \centering
    \resizebox{0.8\textwidth}{!}{ % 将宽度缩放到页面宽度的80%
        \begin{tabular}{c|c|c|c|c|c|c|c}  
            \hline
            Model & \begin{tabular}[c]{@{}c@{}}Frame Size\\ (H$\times$W)\end{tabular} & \begin{tabular}[c]{@{}c@{}}Patch Size\\ (H$\times$W)\end{tabular} & Depth & Hidden & Head & \begin{tabular}[c]{@{}c@{}}Param\\ (M)\end{tabular} & \begin{tabular}[c]{@{}c@{}}Flops\\ (T)\end{tabular} \\ 
            \hline
            LCR-S & 112$\times$112   & 8$\times$8    & 1  & 192   & 4   & 0.82   & 0.003  \\ 
            LCR   & 112$\times$112   & 8$\times$8    & 2  & 384   & 4   & 5.14   & 0.022  \\ 
            LCR-L & 112$\times$112   & 8$\times$8    & 4  & 384   & 4   & 9.97   & 0.041  \\ 
            \hline
            LCR    & 224$\times$224   & 16$\times$16  & 4  & 768   & 12   &37.62  & 0.16  \\ 
            LCR-L  & 224$\times$224   & 16$\times$16  & 8  & 768   & 12   &74.62  & 0.31  \\ 
            LCR-XL & 224$\times$224   & 16$\times$16  & 12  & 768  & 12  &111.63 & 0.47  \\ 
            \hline
        \end{tabular}
    }
    \caption{\textbf{Model variants} For the Jaster dataset, since the input frame size is 112 $\times$ 112, three types of models with 8 $\times$ 8 patch size are used. For the K400 and SSv2 datasets, the frame size expands to 224 $\times$ 224.}
    \label{tab:model_configs}
\end{table}
\textbf{Inference} For the Kinetics-400 and Something-Something V2 datasets, we follow the pipeline outlined in \cite{fan2021multiscale}, sampling T random frames uniformly from each video. patially, we scale the shorter side to 256 pixels and take three crops of size 224 $\times$ 224 to cover the longer spatial axis. Temporally, we uniformly sample the long video into N clips and average the scores from the last one-third of the frames during evaluation. The score for each test sample is derived from the average of these 3 $\times$ N predictions, with the highest score taken as the final prediction. In our work, each prediction is treated as a single "view."For the Jester dataset, we pad shorter videos and randomly sample longer videos to the same length (T). Spatially, we resize each frame to 112 $\times$ 112 pixels without additional transformations, taking the highest prediction score from the last 10 frames as the final prediction. It is important to note that for the inference time of LCR-XL (64 $\times$ 3 $\times$ 3) reported in Table 2, we utilize 3 temporal clips with 3 spatial crops, resulting in a total of 9 views.
\subsection{Performance Evaluation}
\textbf{Jester} Table \textcolor{red}{4} shows the performance comparison against
 vanilla methods on the Jester dataset.As the results suggest, our best method can achieve 90.83$\%$ of Top-1 accuracy with less parameters(5.14M) and Flops (0.022Tflops),while TimeSformer and the best CNN model are 89.94$\%$(2.37$\%$$\downarrow$)and 90.75$\%$(1.56$\%$$\downarrow$), with 46.6M and 4.8M in parameters,1.568G and 1.346G in flops, respectively. Noted that all models are trained from scratch
 \begin{table}[H]
    \centering
    \resizebox{1\textwidth}{!}{
    \begin{tabular}{c|c|c|c|c|c} \hline
    Methods    & \begin{tabular}[c]{@{}c@{}}Top-1\\ (\%)\end{tabular} & \begin{tabular}[c]{@{}c@{}}Top-5\\ (\%)\end{tabular} & \begin{tabular}[c]{@{}c@{}}Param\\ (M)\end{tabular} & \begin{tabular}[c]{@{}c@{}}Flops\\ (T)\end{tabular} & \begin{tabular}[c]{@{}c@{}}Mem\\ (G)\end{tabular} \\ \hline
    ConvLSTM                 & 82.76  & 94.23  & 7.6   & 59.2  & 2.37     \\ 
    TSN                      & 83.90  & 99.60  & 10.7  & 16    & N/A     \\ 
    MobileNet-Small$^{\dag}$ & 84.69  & 98.70  & 2.30  & 0.42  & 1.90    \\ 
    ResNet3D-10$^{\ast}$     & 88.81  & 99.01  & 14.4  & 18.2  & 1.96     \\ 
    R(2+1)D-RGB$^{\ast}$     & 89.08  & 98.76  & 63.6  & 16.9  & 1.93     \\ 
    MobileNet-Large$^{\dag}$ & 89.40  & 99.11  & 15.8  & 1.98  & 13.1     \\ 
    TimeSformer-L$^{\ast}$   & 89.94  & 99.52  & 4.8   & 43.1  & 2.08    \\
    ResNet3D-18$^{\ast}$     & 89.96  & 99.76  & 33.3  & 34.6  & 2.59    \\
    ResNet3D-50$^{\ast}$     & 90.75  & 99.52  & 46.6  & 50.2  &2.68     \\
    SE-ResNet3D$^{\ast}$,32$\times$3  & 90.64  & 99.84 & 48.7  & 52.3 & 2.68 \\ \hline
    RViT-XL,32$\times$3      & 92.31  & 99.87  & 2.27  & 14.1  & 1.76     \\ \hline
    LCR                   & \textbf{90.83}  &being re-calculated  &5.14& 0.022&beingre-calculated   \\ \hline
    \end{tabular}
    }
    \caption{\textbf{Performance comparison on Jester.} We evaluation the gigabyte memory consumption in a single "view". $({\dag})$ indicates the MobileNet accompany with the LSTM unit.}
    \label{tab:Jester}

    \end{table}
%  \subsection{Ablation Studies}
%  We use the Jeste dataset for our extensive ablation study in this section. The purpose of the ablation study is to demonstrate the following hypotheses:

\section{Conclusion}

%-------------------------------------------------------------------------
In this work, LCR is proposed for video understanding tasks. Specifically, the use of linear complexity CrossRWKV gates in the LCR unit enables the integration of edge information and past time information to improve attention to the subject. The edge information is fed as a cue into the forgetting gate of the LSTM to help the network hope for background features. The interference of redundant information in the video is reduced by a channel masking strategy. This design reduces the spatial complexity and computational complexity.

We also evaluated our method on various public benchmark datasets. The results suggest that excellent performance can be achieved on video action recognition tasks with less GPU memory.

Due to the use of the classical lstm structure, the model's parallel computation capability is not as good as that of CNN and RNN and suffers from the problem of gradient vanishing and gradient explosion. Future exploration of this model will allow it to be scaled up to larger network models.Video prediction and video generation tasks will also be explored based on the proposed

\bibliographystyle{unsrtnat}
\bibliography{references}


\end{document}